\documentclass[11pt,a4paper]{article}
\usepackage[hyperref]{emnlp2020}
\usepackage{times}
\usepackage{latexsym}

\usepackage{graphicx}
\usepackage{amsmath}
\usepackage{amsfonts}
\usepackage{amssymb}
\usepackage{multirow}
\usepackage{url}
\usepackage{graphicx}
\usepackage{svg}
\usepackage{tipa}
\usepackage{booktabs}

\usepackage[disable]{todonotes} 

\usepackage{microtype}

\aclfinalcopy 
\def\aclpaperid{2675} 

\title{Tackling the Low-resource Challenge for Canonical Segmentation}

\author{\parbox{\linewidth}{\centering
Manuel Mager{\rm\affmark[1]}, \"{O}zlem \c{C}etino\u{g}lu{\rm\affmark[1]} \and Katharina Kann{\rm\affmark[2]}} \vspace{.12cm}
\\
\affaddr{\affmark[1]Institute for Natural Language Processing,\\ 
University of Stuttgart, Germany}\\
\affaddr{\affmark[2]University of Colorado Boulder, USA }\\
\affaddr{\texttt{\{manuel.mager, ozlem\}@ims.uni-stuttgart.de}},\\
\affaddr{\texttt{katharina.kann@colorado.edu}}} 
\newcommand*{\affaddr}[1]{#1} 
\newcommand*{\affmark}[1][*]{\textsuperscript{#1}}

\date{}

\begin{document}
\maketitle
\begin{abstract}
Canonical morphological segmentation consists of dividing words into their standardized morphemes. 
Here, we are interested in approaches for the task when training data is limited. We compare model performance in a simulated low-resource setting  for the high-resource languages German, English, and Indonesian to experiments on new datasets for the truly low-resource languages Popoluca and Tepehua.
We explore two new models for the task, borrowing from the closely related area of morphological generation: an LSTM pointer-generator and a sequence-to-sequence model with hard monotonic attention trained with imitation learning. 
We find that, in the low-resource setting, the novel approaches outperform existing ones on all languages by up to $11.4\%$ accuracy. However, while accuracy in \textit{emulated} low-resource scenarios is over $50\%$ for all languages,
for the \textit{truly} low-resource languages Popoluca and Tepehua, our best model only obtains 37.4\% and 28.4\% accuracy, respectively. Thus, we conclude that canonical segmentation is still a challenging task for low-resource languages.
\end{abstract}

\section{Introduction}
\label{sec:introduction}
\begin{figure}
    \centering
    \includegraphics[width=.94\columnwidth]{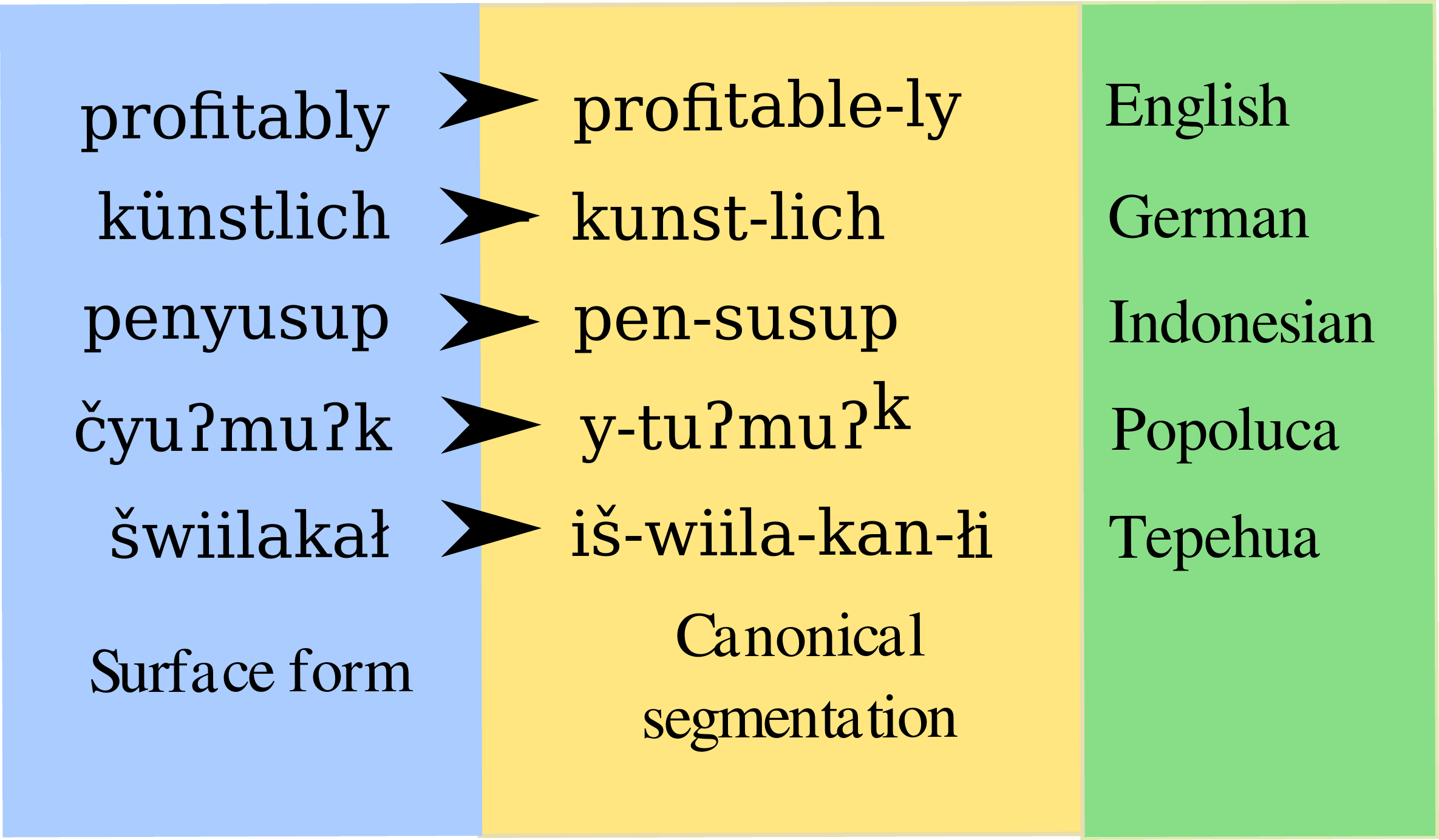}
    \caption{Canonical segmentation examples for all languages in our experiments.}
    \label{fig:examples}
\end{figure}

Morphological segmentation denotes the task of dividing words into their constituting morphemes, i.e., their smallest meaning-bearing units, and has been studied extensively in natural language processing (NLP) \cite{ruokolainen2016comparative}. The most common form of segmentation consists of separating morphemes at the surface level. However, this is not always well suited: in fusional languages, morphemes are merged during word formation and, thereby, change their surface forms. 
Thus, in this paper, we tackle the task of canonical segmentation \cite{cotterell2016joint}, which consists of segmenting a word while restoring the original forms of its morphemes. 
Considering, e.g., the English word \emph{collision}, its surface segmentation is \emph{collis+ion}, while its canonical segmentation is \emph{collide+ion}. Figure \ref{fig:examples} provides examples for all five languages we experiment on.

{}

Neural models have shown to perform well on this task when large amounts of training data are available \cite{kann2016neural,ruzsics2017neural}. Nevertheless, datasets with morphological annotations are difficult to obtain, since they require expert annotators. 
Furthermore, many languages with complex morphology are spoken by a limited number of people or are listed as endangered languages \cite{mager2018b}, which reduces the possible annotator pool even more. 
However, morphological segmentation is important for downstream tasks like machine translation \cite{conforti2018neural, vania2017characters}, dependency parsing \cite{seeker2015graph, vania2018character}, or semantic role labeling \cite{sahin2018character}. Moreover, high performance on these 
tasks can yield more language independent NLP models \cite{gerz2018relation}.
 
Here, we focus on low-resource canonical segmentation. We propose two new models for the task, which have recently been successfully applied to a related morphological generation task called \textit{morphological inflection}. The approaches we investigate are (i) an LSTM pointer-generator model \cite{sharma-etal-2018-iit}, and (ii) 
a neural transducer trained with imitation learning  \citep[IL;][]{makarov-clematide-2018-imitation}.
Since both canonical segmentation and morphological inflection are character-level string transduction tasks, we hypothesize that models which can learn one from limited data, will also be able to do so for the other.
 
We experiment on three benchmark datasets in German, English, and Indonesian, but simulate a low-resource scenario by reducing the number of training examples. We further 
evaluate our models on datasets for two \textit{truly} low-resource languages: Popoluca and Tepehua. We find that our new models indeed outperform previous approaches on all languages. 
For additional insight, we also evaluate the performance of all models for varying amounts of training data from the high-resource languages and find that the neural-transducer with imitation learning outperforms all other models in all but one setting with up to 600 training examples.
Using the entire training set for English, German, and Indonesian, the state-of-the-art LSTM sequence-to-sequence model performs best. However, the difference to our proposed models is below 3.3\% accuracy for all languages and models. 

\paragraph{Contributions.} (i) Inspired by recent advances in the area of morphological generation, we propose two new models for the task of low-resource canonical segmentation, which outperform all baselines. (ii) We introduce two canonical segmentation datasets for the \textit{truly} low-resource languages Popoluca and Tepehua. (iii) We compare all models under multiple different conditions, highlighting their strengths and shortcomings, and conduct an analysis of the errors made by all neural models.

\section{Related Work}
\label{sec:related_work}
The task of morphological segmentation was introduced by \newcite{harris1951methods}. Most work has considered the surface segmentation task, for which unsupervised methods like LINGUISTICA \cite{Goldsmith:2001:ULM:972667.972668} and MORFESSOR \cite{creutz2002, creutz2007unsupervised, poon2009unsupervised} played an important role. The latter was further extended to a semi-supervised version \cite{kohonen2010semi, gronroos2014morfessor}. 

Over the last years, supervised methods have attracted more attention: \newcite{ruokolainen2013supervised} cast the task as a sequence labeling problem using conditional random fields \citep[CRFs;][]{lafferty2001conditional}. A similar approach was suggested by \newcite{wang2016morphological}, who employed a long short-term memory network \citep[LSTM;][]{hochreiter1997long} for tagging. Semi-Markov CRFs were also proposed \cite{cotterell2015labeled}. 
\newcite{kann2018fortification} modeled the task as a sequence-to-sequence problem. Supervised methods for surface segmentation were shown to perform acceptably even in the low-resource setting \cite{gronroos2019north}. 
Recent work also included context to improve morphological disambiguation \cite{burcu2018, tarek2017}. 
\newcite{yang2019point} proposed a pointer network to find surface segmentation boundaries.

For fusional languages, surface segmentation is not very effective.
Therefore, restoring morphemes to their canonical form was previously discussed in linguistics \cite{kay1977morphological} as well as in the NLP literature. Previous approaches include unsupervised  \cite{naradowsky2009improving}, as well as joint models for segmentation and transduction \cite{cotterell2016joint} and neural encoder-decoder models \cite{kann2016neural, ruzsics2017neural}. However, up to now, supervised models have
only been explored in the high-resource setting. We aim at closing this gap.

For low-resource morphological segmentation, rule-based approaches have been used frequently, since they do not need large amounts of data. They have been developed, e.g., with finite state transducer (FST) tools
like FOMA \cite{hulden2009foma} or HFST \cite{linden2011hfst}. 
However, this kind of system requires both time and linguistic knowledge. Our aim is to explore data-driven approaches for the low-resource setting in order to overcome this limitation. 

In recent years, the area of morphological generation has experienced substantial progress, with a variety of methods that can be used for the canonical segmentation task. \newcite{kann2016neural} used a sequence-to-sequence model to inflect a word given a set of morphological tags. \citet{sharma-etal-2018-iit} proposed a pointer-generator model, which was more suitable for the low-resource setting. \newcite{aharoni-goldberg-2017-morphological} proposed a neural transducer with hard monotonic attention. \citet{makarov2017align} extended this approach and added a copy operation, and \newcite{makarov-clematide-2018-imitation} proposed  imitation learning \cite{daume2009search} for training it. 
Here, we explore the applicability of the models by \citet{sharma-etal-2018-iit} and \newcite{makarov-clematide-2018-imitation} to low-resource canonical segmentation.

\section{Datasets for Popoluca and Tepehua}
We release two new datasets for low-resource canonical segmentation in Popoluca and Tepehua\footnote{Te dataset is available at \url{http://turing.iimas.unam.mx/wix/canseg}}. In this section, we briefly introduce the languages, before describing our datasets. We use these two languages to shed light on polysynthetic languages that also exhibit fusional phenomena. The high-resource datasets introduced by \cite{cotterell-etal-2016-morphological} cover fusional (German), analytic (English), and agglutinative (Indonesian) languages.

\subsection{Languages}

In addition to experimenting on high-resource datasets for English, German and Indonesian \cite{cotterell2016joint}, we introduce datasets for two low-resource languages from Mexico: Popoluca and Tepehua. This enables us to evaluate our models in real low-resource settings.
\begin{table}[]
    \centering
    \small
    \setlength{\tabcolsep}{2.5pt}
    \begin{tabular}{l r|l r|l r|l r|l r}
    \toprule
\multicolumn{2}{c|}{ENG} & \multicolumn{2}{c|}{DEU} &\multicolumn{2}{c|}{IND} &\multicolumn{2}{c|}{POQ} &\multicolumn{2}{c}{TTP} \\\midrule
ly & 7.53 & er & 15.66 & men & 8.65 & y & 6.08 & ya & 8.58 \\ 
ness & 3.41 & in & 10.38 & nya & 8.29 & $\emptyset$ & 6.08 & \l{}i & 6.27 \\ 
er & 2.99 & ung & 8.14 & an & 7.18 & n & 3.98 & ka & 4.46 \\
ion & 1.87 & lich & 4.37 & kan & 6.61 & ny & 3.56 & ta & 4.29 \\ 
y & 1.50 & keit & 3.96 & di & 5.31 & k & 3.35 & ti & 3.80 \\
ity & 1.24 & ig & 3.78 & pen & 4.14 & p & 2.94 & ik & 2.81 \\ 
ation & 0.99 & los & 1.23 & ber & 2.81 & t+k & 2.52 & ni & 2.64 \\ 
un & 0.88 & chen & 1.16 & i & 2.45 & ky & 2.31 & \v{c}a & 2.31 \\
ic & 0.85 & bar & 1.13 & ter & 1.91 & wat & 2.10 & la & 1.82 \\ 
al & 0.81 & ver & 0.81 & per & 1.25 & a\textglotstop & 2.10 & maa & 1.82 \\
ist & 0.76 & un & 0.77 & se & 0.72 & ta\textglotstop & 1.89 & kin & 1.82 \\ 
able & 0.74 & e & 0.49 & ke & 0.71 & \textglotstop e\v{s}  & 1.26 & waa & 1.82\\ 
\bottomrule
          
    \end{tabular}
    \caption{Relative frequencies of the 12 most common morphemes for each language; ENG=English; DEU=German; IND=Indonesian; POQ=Popoluca; TTP=Tepehua.}
    \label{tab:morphs}
\end{table}

\paragraph{Popoluca. } Popoluca of Texistepec (language code: POQ\footnote{We use the languages codes defined in the ISO 639-3 standard.})  
is part of the Mixe-Zoquean family. 
Its morphology is classified as polysynthetic, and it mostly follows a verb, subject, object (VSO) word order \cite{wals}. This language is almost extinct with only one native speaker alive reported in 2005 \cite{gordon2005ethnologue}. 
However, attempts for language revival have been reported \cite{instituto2008programa}.
Efforts made for language revitalization can benefit from advances in NLP. Thus, the creation and development of accurate models for those languages is of high importance.

Here we show an example of canonical segmentation in Popoluca, together with its English gloss. The plus symbol is part of the alphabet of the language. We use a `-' as morpheme delimiter.  
\begin{center}
kki:?mba: $\to$ ky-k+:?m-ba: 

\textit{You are small}
\end{center}

\paragraph{Tepehua. } Tepehua (language code: TPP) belongs to the Totonacan language family. It is spoken in three Mexican regions: in the northeastern part of the state of Hidalgo (around 3000 speakers), in the villages of Pisaflores (around 4000 speakers), and in Tlachichilco in the state of Veracruz (around 3000 speakers) \cite{gordon2005ethnologue}. It is also polysynthetic. 
Tepehua permits free word order, but has a preference for a subject, verb, object (SVO) configuration \cite{wals}.  

An example for canonical segmentation is
\begin{center}
iklakadíkdi $\to$ ik-laka-tikti

\textit{I am small}
\end{center}

The variant of the language used in our dataset is the one spoken in Pisaflores, Veracruz. 
\begin{table}[!t]
    \centering
    \small
    \setlength{\tabcolsep}{2.5pt}
    \begin{tabular}{c|r|r|r|r|r|r}
        \toprule
          & $>$3Morph. & Surf. & Canon. & NoSeg. & M./W. & Ch./W. \\\midrule
        ENG   & 00.01 & 36.40 & 22.83 & 41.37 & 01.60 &  08.18 \\
        DEU    & 01.86 & 46.07 & 53.86 & 00.00 & 02.20 & 12.48 \\
        IND & 05.57 & 46.21 & 23.66 & 30.14 & 02.07 &  08.65\\
        POQ  & 12.12 & 23.74 & 56.57 & 19.70 & 02.41  &  06.78\\
        TTP   & 32.00 & 21.50 & 63.00 & 15.50 & 03.03 &  08.62\\\bottomrule
        
    \end{tabular}
    \caption{Statistics for all five canonical segmentation datasets. Percentages of words with more than 3 morphemes ($>$3 Morph.), surface segmentation (Surf.), canonical segmentation (Canon.), and without segmentation (NoSeg.), as well as the average number of morphemes per word (M./W.) and characters per word (Ch./W.).}
    \label{tab:dataset_stats}
\end{table}{}
\subsection{Datasets}

We collect words for our datasets from two books belonging to the Archive of Indigenous Languages (ALI-Colmex) of the College of Mexico (\textit{Colegio de M\'exico}). 
For Popoluca we used the book by \newcite{wichmann2007popoluca} and for Tepehua that by \newcite{mackay2010tepehua}. We include segmentable as well as non-segmentable words in order to avoid oversegmentation by our systems. 
For both languages a set of Spanish sentences are used to elicit the data. This set of sentences is the same across the entire ALI-Colmex collection. For each language the authors of the books asked native speakers to translate the sentences into the respective languages (elicited data). Afterwards, they performed a glossing of the translated text. For more details we refer the reader to the original books. 

In Table \ref{tab:dataset_stats}, we show statistics for all five datasets used in this paper. Importantly, the German dataset only contains multi-morpheme words. Additionally, we observe that most of the Indonesian words only require surface segmentation, while English is the language with the highest ratio of words that do not require any segmentation. On the other hand, Popoluca and Tepehua have the highest proportion of words that require both splitting and restoration of the canonical forms. Moreover, both languages have a high amount of words that contain more than 3 morphemes per word, and also have the highest morphemes-per-word rate. Adding to these facts, the small amount of data available for these languages makes morphological segmentation even harder. To get a better understanding of the underlying morphemes seen in each language, we extract the 15 most common ones for each dataset. These morphemes, together 
with their relative frequency in our datasets, are shown in Table \ref{tab:morphs}.

\section{Models}
\label{sec:models}
Inspired by recent successes of two models for low-resource morphological inflection, we propose to apply these architectures to canonical segmentation with limited training data. In this section, we introduce the models.

\subsection{Pointer-Generator Network}
\paragraph{Motivation.} The first model we apply to low-resource canonical segmentation is a pointer-generator network \cite{P17-1099}, i.e., a sequence-to-sequence model with a mechanism to copy input elements over to the output. Our intuition is that this should make the learning problem easier and 
help in settings with limited training data.
The pointer-generator network can be considered a hybrid between an attention-based sequence-to-sequence model \cite{bahdanau2015neural} and a pointer network \cite{vinyals2015pointer}.

\paragraph{Model description. }

Our pointer-generator network consists of a bidirectional LSTM \cite{hochreiter1997long} encoder and a unidirectional LSTM decoder with an attention mechanism.
We cast the task of canonical segmentation as a character-based sequence-to-sequence problem, with the characters of the original word as the input and the characters of the restored morphemes in combination with segment boundary markers as the output. Both our encoder and decoder operate on the character level.

The pointer-generator network differs from the standard sequence-to-sequence architecture in that the decoder calculates a probability for copying an element from the input over to the output instead of generating.
Here, we follow \citet{sharma2018iit} and use two separate encoders: one for the lemma and one for the morphological tags. The decoder then computes the probability distribution of the output at each time step as a weighted sum of the probability distribution over the output vocabulary and the attention distribution over the input characters. The weights can be seen as the probability to generate or copy, respectively, and are computed by a feed-forward network.
For details, we refer the reader to \newcite{sharma2018iit}.

\paragraph{Hyperparameters.} All encoder and decoder hidden states are $100$-dimensional, and our embeddings are of size $100$. For training, we use Adam \cite{kingma2014adam} with a learning rate of $0.001$ and a mini-batch size of $32$. To avoid overfitting, we use dropout \cite{srivastava2014dropout} with a coefficient of $0.3$ for the high-resource setting and $0.5$ for the low-resource setting. We train our model for $100$ and $300$ epochs and use early stopping with a patience of $10$ and $100$ for the high-resource and the low-resource setting, respectively.  

\subsection{Neural Transducer with Imitation Learning}
\paragraph{Motivation.}

Hard monotonic attention networks
 \cite{aharoni-goldberg-2017-morphological} have shown to perform well on morphological generation in the low-resource setting.  These systems use a nearly-monotonic alignment between the source characters and the output characters. For our second model, we employ the variant proposed by \citet{makarov-clematide-2018-neural}, which makes use of imitation learning for end-to-end training and, thus, avoids error propagation.

\paragraph{Model description.}

This model is a sequence-to-sequence model with hard monotonic attention \cite{aharoni-goldberg-2017-morphological}, which transduces an input sequence of characters into an output sequence by performing edit operations. Following \citet{makarov2018imitation}, it can perform three operations: insertion, deletion and copy. However, instead of using maximum likelihood estimation (MLE), training is done with imitation learning. The idea is to train a model to imitate an expert policy that maps the training configurations to a set of optimal actions. We aim to minimize the sequence-level loss and an action level loss.

The training is composed of two steps: a roll-in and a roll-out stage. In the roll-in stage, the model gather actions by sampling from the expert policy. This process returns a set of decoder outputs called configurations. For the roll-out stage: a sequence-level loss is computed for each valid action per configuration. For that, the action is executed and is compared to the optimal action sequence of the expert. This loss is defined in terms of Levenshtein distance \cite{levenshtein1966binary} between the prediction and the target and the cost of the actions. The cost function uses the information from a character aligner. After calculating the sentence-level loss, this is fed into an action-level loss. This loss expresses how much a certain action suffers relative to the optimal action under the current policy.  This is done by minimizing the negative marginal log-likelihood of all optimal actions \cite{makarov2018imitation}. 

\paragraph{Hyperparameters.}

For the encoder and the decoder of this model, we use one layer with a 200-dimensional size, with a dropout of 0.5.  For optimization we use ADADELTA \cite{zeiler2012adadelta} with a learning rate of 0.1. As the RNN unit, we use an LSTM.  We train the model for 30 epochs, with a patience of 10 epochs. For IL training, we use an inverse sigmoid, and a decay rate of 12. For decoding, we employ beam  search with a beam of width 4. 

\section{Experiments}
\label{sec:experiments}
We now describe the experiments we conduct to explore the performance of our models both in the high-resource and in the low-resource setting.

\begin{table*}[!htbp]
    \centering
    \setlength{\tabcolsep}{5pt}
    \small
\begin{tabular}{ l |  r | r | r|  r | r | r |  r | r | r  }
\toprule
             & \multicolumn{3}{c |}{English} &  \multicolumn{3}{c |}{German} & \multicolumn{3}{c }{Indonesian}   \\\midrule
                                  &  Acc.& ED & F1 &  Acc.& ED & F1 & Acc.& ED & F1 \\\midrule
             \texttt{SemiCRF}     &    64.7 &     64.3 &     76.6 &    41.9 &    108.3 &     74.1 &     70.4 &     46.3 &     84.3 \\
             \texttt{joint}       &    72.0 &     98.0 &     76.0 &    59.0 &    101.0 &     76.0 &     90.0 &     15.0 &     80.0 \\
             \texttt{s2s}         &\bf $^\blacklozenge78.0$ & \bf 41.2 &     88.4 &\bf $^\blacklozenge77.1$ & \bf 47.8 & \bf 89.3 & \bf $^\blacklozenge94.3$ & \bf  7.6 & \bf 97.9 \\
             \texttt{PGNet}       &    77.5 &     42.4 & \bf 88.5 &    74.8 &     52.1 &     88.2 &     92.9 &     10.0 &     97.5 \\
             \texttt{IL}          &    76.7 &     42.9 &     87.2 &    73.8 &     52.3 &     87.2 &     93.4 &      8.4 &     97.6 \\
             \bottomrule
             
    \end{tabular}
    \caption{Results for \texttt{semiCRF}, \texttt{joint}, \texttt{s2s}, \texttt{PGNet}, and \texttt{IL} for the high-resource setting of English, German and Indonesian. Lower scores in the ED columns are better. For accuracy, $^\blacklozenge$ indicates statistical significance at $p < .01$.}
    \label{tab:results}
\end{table*}

\subsection{Data}
\label{sec:data}
The canonical segmentation datasets for English (ENG), German (DEU) and Indonesian (IND)  by \newcite{cotterell2016joint} each consist of $8000$ training, $1000$ development, and $1000$ test examples. We consider the complete training set to be high-resource. The datasets feature a splitting into 10 folds for cross-validation. For our low-resource experiments, we randomly take a subset of  
words from each training fold, but keep the development and test sets unchanged. 

The high-resource datasets cover three languages: English, German, and Indonesian. 
English is an analytic language from the Indo-European family \cite{konig2013germanic}, German exhibits fusional typology \cite{hawkins2015comparative}, while Indonesian is an agglutinative language whose morphology involves the use of affixation, reduplication and cliticization \cite{NOMOTO18.8}.

We additionally experiment with two polysynthetic low-resource languages: Tepehua and Popoluca (cf. Section \ref{tab:dataset_stats}). As those datasets are small (900 words for each language), we divide the datasets into 9 folds, each containing 100 training, 100 development, and 700 test examples.

\subsection{Baselines}
\label{sec:baselines}

We compare the neural-transducer with imitation-learning (\texttt{IL}) and the pointer-generator network (\texttt{PGNet}) to three strong baselines, including the current state of the art for the canonical segmentation task. 

\paragraph{Encoder-Decoder (\texttt{s2s}).} Our first baseline is a character-based encoder-decoder recurrent neural network (RNN) architecture with attention as proposed by \newcite{kann2016neural}. It defines (in combination with a reranker which we omit here since it is orthogonal to our work) the state of the art on the high-resource datasets. To perform experiments in the low-resource setting, we re-implement this model using OpenNMT \cite{klein2017opennmt}. The hyperparameters suggested by \newcite{kann2016neural} are as follows: the  RNNs of the encoder and decoder have 100 hidden units each; the embedding size is 300. For optimization we use ADADELTA \cite{zeiler2012adadelta} with a minibatch size of 20.

\paragraph{Semi-Markov CRF (\texttt{semiCRF}).} Our first non-neural baseline is the ChipMunk \cite{cotterell2015labeled} implementation of a semi-Markov CRF \cite{sarawagi2005semi}. Although the system is able to make use of additional complementary information like morphological tags or dictionaries, we decide to not include those, in order to make our results comparable across all languages and systems. 

\paragraph{Joint log-linear model (\texttt{joint})} As a second non-neural system we use a log-linear model which jointly segments and generates underlying representations of the input words \cite{cotterell2016joint}. For segmentation it uses the \texttt{semiCRF} previously described, and for transduction of the underlying forms it uses a probabilistic final state transducer \cite{cotterell2014stochastic}.

\subsection{Training Details}

We choose the hyperparameters for all models following the mentioned previous work. All neural models and the \texttt{semiCRF} were trained on a server with 2 Intel(R) Xeon(R) CPU v4@ 2.20GHz, with 4 Nvidia GTX 1080ti graphic cards. To train the joint log-linear model a MacBook Pro 2009 laptop was used. Links to the repositories we use are listed in the complementary material.

\begin{table*}[htbp!]
    \centering
    \small
    \setlength{\tabcolsep}{5.5pt}
    \begin{tabular}{l|r|r|r|r|r|r}
    \toprule
                  & \multicolumn{3}{c|}{Tepehua} & \multicolumn{3}{c}{Popoluca} \\
                  \midrule
         Model & Acc. & ED & F1 & Acc. & ED & F1 \\
                  \midrule
        \texttt{SemiCRF}     &     21.9 &     285.3 &      35.9 &     26.0 &    215.0 &     41.4 \\
        \texttt{joint}       &     11.2 &     335.4 &      29.5 &     14.6 &    393.6 &     36.8 \\
        \texttt{s2s}     &      4.1 &     532.4 &       7.7 &     13.2 &    309.4 &     23.3 \\
        \texttt{PGNet}       &     17.2 &     321.7 &      29.3 &     27.0 &    211.0 &     42.5 \\
        \texttt{IL}          & \bf $^\blacklozenge28.4$ & \bf 242.6 & \bf  44.0 & \bf $^\blacklozenge37.4$ &\bf 158.8 & \bf 54.7 \\\bottomrule
    \end{tabular}
    \caption{Results for the low-resource languages Popoluca and Tepehua. For accuracy, $^\blacklozenge$ indicates statistical significance at $p < .01$.
    }
    \label{tab:minimal}
\end{table*}
\subsection{Metrics}
\label{sec:metrics}

For evaluation, we use three metrics. The first one is \textbf{accuracy}, i.e., the proportion of entirely correctly segmented words, to get a better understanding of partially right segmentation. To get more information about subword-level errors, we also employ \textbf{edit distance} on the character level. This is particularly useful to penalize big mistakes in a single word. We also use $\textbf{F}_1$ \textbf{score} on the morpheme level, to measure the overlap between morphemes. Precision corresponds to the proportion of morphemes in the prediction that occur in the gold standard, and recall is the proportion of morphemes in gold that appear in the system's prediction. This will ensure that morphemes that are predicted without appearing in the gold standard are penalized, as well morphemes that are in the gold standard but are omitted in the prediction.

\subsection{Results}

\begin{figure*}
    \centering
    \includegraphics[width=0.9\textwidth]{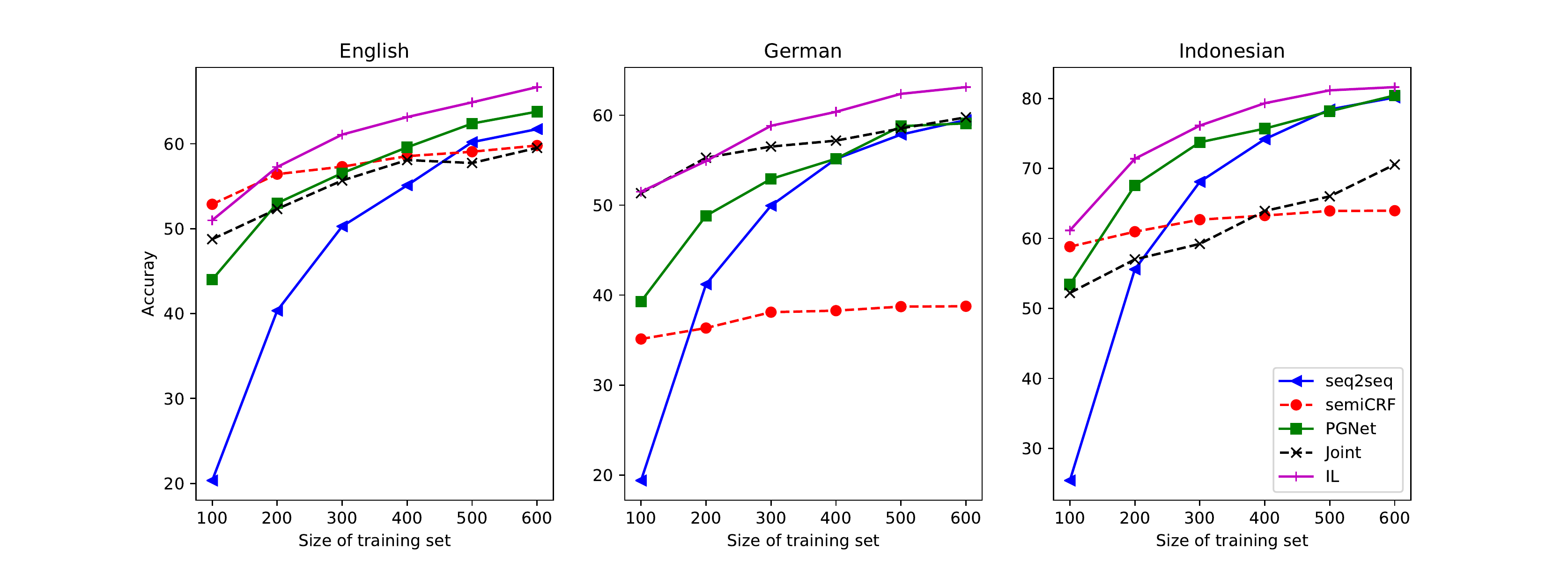}
    \caption{Accuracy for different simulated low-resource settings for our high-resource languages. }
    \label{fig:results_steps}
\end{figure*}
\paragraph{Low-resource \textit{simulation}.}  
Figure \ref{fig:results_steps} shows the accuracy of all systems for  
different low-resource training set sizes (100, 200, 300, 400, 500 and 600 examples) for English, German, and Indonesian. To ensure statistical significance we use McNemar's test \cite{McNemar1947} for all accuracy results (Tables \ref{tab:results} and \ref{tab:minimal}, Figure \ref{fig:results_steps}) comparing the best and the second best systems. All results are significant at $p< 0.01$. The scores of all systems vary across languages. However, \texttt{IL} consistently is among the two best systems in terms of accuracy in all settings. For 100 training examples \texttt{IL} is  the second-best performing system with $50.99\%$ for English, just behind the \texttt{semiCRF} with $52.87\%$. For German $51.49\%$ \texttt{IL} slightly outperforms \texttt{Joint} ($51.33\%$) and obtains the best score for Indonesian with $61.14\%$, where the second best system is \texttt{semiCRF} ($58.82\%$). Moreover, from 300 examples up to 600, \texttt{IL} strongly outperforms all other systems, including non-neural ones.

If we compare the performance of our two proposed systems with \texttt{s2s}, \texttt{PGNet} strongly outperforms \texttt{s2s} with improvements of $22.27\%$, $17.84\%$, and $25.66\%$ absolute accuracy for English, German, and Indonesian, respectively, in the setting with 100 training examples; while \texttt{IL} have even bigger gains with improvements of $30.65\%$, $32.1\%$ and $35.73\%$ of accuracy respectively.

Looking on the learning curves for each model for increasing training set sizes, we can see that both proposed systems show monotonically increasing performance: they take advantage of more data well, but still achieve decent performance in the low-resource setting, even outperforming all non-neural systems in some settings.  
On the contrary, the non-neural models \texttt{joint} and \texttt{semiCRF} have in many cases a good start, but only benefit to a limited extends from additional data.
A table listing all individual results for this experiment is included in the supplementary material.

\paragraph{Low-resource \textit{languages}.} Results for Popoluca and Tepehua are shown in Table \ref{tab:minimal} and confirm most of the tendencies seen in our low-resource simulation experiment. \texttt{s2s} barely predicts any correct segmentation for Tepehua, and only obtains $4.14\%$  
absolute accuracy and $13.19$ $F_1$ score. Similarly, for Popoluca, \texttt{s2s} reaches only $13.19\%$ accuracy. 
The performance of \texttt{IL} is consistently better on all metrics, with substantial gains for Tepehua of $6.5\%$ accuracy over the closest system (\texttt{semiCRF}) and $10.4\%$ accuracy over \texttt{PGNet}. 

The performance of \texttt{PGNet} is consistently better than that of \texttt{s2s}, with gains of $13.03\%$ and $13.77\%$ accuracy for Tepehua and Popoluca, respectively.
\texttt{joint} surprisingly shows a low performance for our two low-resource languages, obtaining a $17.2\%$ lower accuracy than the best model for Tepehua (\texttt{IL}), and a $22.8\%$ lower accuracy than the best system for Popoluca (\texttt{IL}). 

Overall, all systems perform notably worse for Tepehua and Popoluca than for the high-resource languages. This could be due to their high morphological complexity, as shown in Table \ref{tab:dataset_stats}.

\paragraph{High-resource setting.} Table \ref{tab:results} shows results for \texttt{IL, }\texttt{PGNet}, \texttt{s2s}, \texttt{joint}, and \texttt{semiCRF} for the high-resource experiment. The \texttt{s2s} model gets the best results in this setting with $78.02\%$, $77.06\%$, and $94.30\%$ accuracy for English, German, and Indonesian, respectively.
However, it only obtains a slightly higher accuracy than \texttt{PGNet}  
and the differences in $F_1$ scores are similarly small. Overall, the pointer-generator network achieves results that are comparable with the state of the art in the high-resource setting. 
In contrast to the good performance for low-resource settings, \texttt{IL} under-performs on all metrics compared to \texttt{s2s} and \texttt{PGNet}.
The \texttt{joint} model is the best non-neural system and performs clearly worse than both neural systems. Compared to \texttt{PGNet}, its accuracy is $5.54\%$ lower for English, $15.80\%$ lower for German, and $2.90\%$ lower for Indonesian. \texttt{semiCRF} performs even worse.

\begin{table}[ht!]
    \centering
    \setlength{\tabcolsep}{5.pt}
    \begin{tabular}{c p{4.5cm}}
         \multicolumn{2}{c}{\bf Oversegmentation}  \\
         Input & internationalisierung \\
         Gold & internationale isier ung\\
         Error & internationale \textit{is i er} ung\\
         Description & The morpheme \texttt{isier} is segmented wrongly into three morphemes.\\\hline
         
         \multicolumn{2}{c}{\bf Undersegmentation} \\
         Input & internationalisierung \\
         Gold & internationale isier ung\\
         Error & internationale \textit{isieung}\\
         Description & The morphemes \texttt{isier} and \texttt{ung} are lacking of a segmentation boundary.\\\hline
         
         \multicolumn{2}{c}{\bf Restoration Error} \\
         Input & internationalisierung \\
         Gold & internationale isier ung\\
         Error & \textit{international} isier ung\\
         Description & The system did not perform the needed restoration for the stem \texttt{internationale}.\\\hline
         
         \multicolumn{2}{c}{\bf Overrestoration}\\
         Input & internationalisierung \\
         Gold & internationale isier ung\\
         Error & internationaler \textit{isierer} ung\\
         Description & The systems perfomed a restoration on a morpheme that is not supposed to be restored.\\\hline
         
         \multicolumn{2}{c}{\bf Wrong segmentation} \\
          Input & internationalisierung \\
         Gold & internationale isier ung\\
         Error & internationale \textit{isi erung}\\
         Description & The segmentation was done with the exact number of morphemes as in gold, however, the segmentation points are wrongly placed. In this error count all instances that do not match the exact segmentation boundaries.\\
         
    \end{tabular}
    \caption{Examples of error types. Wrong parts are marked in italics.}
    \label{tab:error_examples}
\end{table}

\section{Error Analysis}
\label{sec:error_analysis}

To get a better understanding of the results obtained with our neural models, we perform an error analysis on the output for the development sets of all folds. By manual inspection, we identify five not mutually exclusive types of errors: 
\textbf{Oversegmentation (Overseg.)} arises when the number of morpheme boundaries in the prediction is higher than in the gold standard annotation. 
\textbf{Undersegmentation (Underseg.)} occurs when the number of morpheme boundaries is lower than in the gold standard.
\textbf{Restoration error (Res.)} occurs when the prediction does not match the gold annotation, and the predicted word without boundaries does not match the input. These are errors that occur to words that undergo orthographic changes during word-formation. 
\textbf{Overrestoration (Overres.)} refers to outputs with errors where the correct output needs only segmentation and a copy of the input to the output. 
\textbf{Wrong segmentation (Wrong seg.)} arises when the morpheme boundaries in the prediction are not the same as in gold. From each segmented word, we extract the indices within the word where the segmentation is performed. If the segmentation indices from the gold standard and the prediction are not equal, it counts as this error.

Table \ref{fig:error} shows the percentage of errors in all languages for both experimental settings (100 examples in the low-resource setting). For the high-resource experiments, the results for oversegmentation and undersegmentation errors are mixed: for English, \texttt{s2s} avoids to generate too many segmentation boundaries, but this also has the drawback of not segmenting sufficient when it is needed. The opposite happens for German, where \texttt{IL} performs better as well, with respect to oversegmentation but fails regarding undersegmentation. 
\texttt{PGNet} shows no strong wins or problems regarding these errors, except for English, where it performs better for undersegmentation.  
\texttt{s2s} performs better for restoration errors with the exception of English, where again \texttt{PGNet} improves. With respect to oversegmentation errors, \texttt{IL} wins on all languages when compared to the other neural systems. As Indonesian has a relatively regular morphology, all error types are much less frequent for this language. 
If we only consider the exact segmentation point prediction, \texttt{s2s} performs better for all languages. However, the differences between the observed error rates are relatively small between \texttt{s2s} and \texttt{PGNet} models. Overall, wrong segmentation errors are the most common error type for all languages in the high-resource setting.

\begin{table*}[!htb]
    \centering
    \setlength{\tabcolsep}{2.pt}
    \small
    \begin{tabular}{c | c | c| c | c | c | c | c | c | c | c | c | c| c | c | c | c}
    \toprule
         & &  \multicolumn{3}{ c | }{ Overseg. } & \multicolumn{3}{ c | }{ Underseg. } & \multicolumn{3}{ c | }{ Res.} & \multicolumn{3}{ c | }{ Overres.} & \multicolumn{3}{c}{Wrong seg.}\\\midrule
         & & \texttt{IL} & \texttt{PGNet} & \texttt{s2s} & \texttt{IL} & \texttt{PGNet}  & \texttt{s2s} &\texttt{IL}& \texttt{PGNet}  & \texttt{s2s} & \texttt{IL}& \texttt{PGNet}  & \texttt{s2s} & \texttt{IL}& \texttt{PGNet}  & \texttt{s2s} \\\midrule
&ENG      &      5.54 & 05.60 &\bf 05.28 &     08.13 & \bf 06.58 &     07.37 &     08.04 & \bf 05.86 &     06.28 & \bf 02.34 & 05.30 &     04.00 &     21.67 &     19.68 & \bf 17.01 \\
High &DEU & \bf  4.17 & 04.42 &    04.88 &     09.30 &     08.11 & \bf 07.02 &     09.24 &     07.42 & \bf 06.94 & \bf 06.08 & 07.55 &     06.40 &     25.46 &     23.65 & \bf 20.49 \\
&IND      &      2.26 & 02.52 &\bf 01.91 &     01.76 &     01.67 & \bf 01.50 &     00.46 &     00.52 & \bf 00.45 & \bf 00.58 & 01.22 &     00.79 &     05.16 &     05.29 & \bf 03.26 \\\midrule
&ENG      & \bf  5.84 & 07.52 &    11.97 &     26.06 & \bf 18.82 &     21.75 &     18.94 &     10.39 & \bf 04.96 & \bf 02.56 & 20.48 &     49.43 & \bf 46.92 &     48.35 &     70.19 \\
&DEU      & \bf  1.40 & 04.11 &    07.79 &     17.56 & \bf 14.83 &     15.70 &     32.01 &     16.26 & \bf 07.81 & \bf 03.94 & 21.88 &     33.93 & \bf 41.66 &     51.52 &     71.78 \\
Low&IND   & \bf 10.94 & 11.03 &    15.47 &     15.24 & \bf 10.61 &     14.00 &      4.91 &     03.19 & \bf 01.46 & \bf 02.96 & 19.90 &     50.00 &     34.64 & \bf 34.25 &     36.16 \\
&TPP      & \bf 15.86 & 27.75 &    34.45 &     42.43 & \bf 07.56 &     23.42 &     32.16 & \bf 07.58 &     14.10 & \bf 03.80 & 25.04 &     44.20 & \bf 69.52 &     73.39 &     86.34 \\
&POQ      & \bf 15.86 & 21.88 &    26.10 &     28.43 & \bf 10.22 &     25.18 &     22.86 & \bf 10.29 &     17.68 & \bf 07.86 & 22.17 &     49.42 & \bf 55.71 &     57.54 &     76.81 \\ \bottomrule

    \end{tabular}
    \caption{Error types found in the development set. The high resource configuration includes three languages, while the low-resourced setting refers to model performance using 100 training examples. This error analysis was done for all five languages.}
    \label{fig:error}
\end{table*}

In the low-resources experiments, \texttt{IL} excels over all other models for oversegmentation and overrestoration, and for all languages with except to Indonesian for wrong segmentation errors. This low error rate explains the important gains that this model shows for low-resource languages. \texttt{PGNet}  shows, however, better performance avoiding undersegmentation errors in all languages. It also performs better for Popoluca and Tepehua for restoration errors, while \texttt{s2s} has the lowest restoration errors for English, German, and Indonesian.

Finally, we also perform an error analysis of \texttt{joint} (cf. supplementary material). In our low-resource simulation experiments, we notice a surprisingly good performance of \texttt{joint} for German. The data for this language is special since all words contained in the set are segmentable. We find that \texttt{joint}  has no undersegmentation errors at all. Also, it makes very few copy errors ($9.5\%$, compared to $21.7\%$ of \texttt{PGNet}). For our new datasets, this model obtains a high rate of wrong segmentation ($88.87\%$ for Popoluca and $91.57\%$ for Tepehua). It further seems to not easily be able to decide which words should or should not be segmented. This is shown by the high undersegmentation rate ($50.14\%$ for Popoluca and $62.43\%$ for Tepehua). Thus, the low performance of \texttt{joint} on those languages can be explained by this error type and the high morphemes-per-word rate of those languages as shown in Table \ref{tab:dataset_stats}.

\section{Conclusion}
We proposed two new models for the task of canonical segmentation in the low-resource setting: an LSTM pointer-generator model and 
a neural transducer trained with imitation learning.
We evaluated the performance of both models against multiple state-of-the-art baselines on five languages of different morphological typology: English, German, Indonesian, Tepehua, and Popoluca. In emulated low-resource settings with 
up to 600 training examples,
our best proposed model
outperformed all baselines in all but one setting.
We obtained a similar picture for experiments on the truly low-resource languages Popoluca and Tepehua: our best approach outperformed the best baseline by $11.4\%$ and $6.5\% $ accuracy. For large training sets, our systems performed close to the state of the art.
However, we find a large gap between the \textit{emulated} and the \textit{real} low-resource scenarios: while accuracy is above $50\%$ for all high-resource languages even with reduced amounts of training data,
for Popoluca and Tepehua, our best model only obtains 37.4\% and 28.4\% accuracy, respectively.

\section*{Acknowledgments}
We would like to thank Hilaria Cruz, Jaime Gonzalez and Ekta Sood for their valuable feedback. We also want to thank all the anonymous reviewers for their valuable comments and suggestions. 
This project has benefited from financial support to Manuel Mager and \"{O}zlem \c{C}etino\u{g}lu by DFG via project CE 326/1-1 ``Computational \underline{S}tructural \underline{A}nalysis of \underline{G}erman-\underline{T}urkish Code-Switching'' and to Manuel Mager by DAAD Doctoral Research Grant.

\bibliographystyle{acl_natbib}
\bibliography{anthology,emnlp2020}

\cleardoublepage
\appendix

\section{Appendices}
\label{sec:appendix}

\begin{table}[ht!]
    \centering
    \footnotesize
    \setlength{\tabcolsep}{3.0pt}
    \begin{tabular}{c|c||r|r|r||r|r|r||r|r|r||r|r|r||r|r|r}
    \toprule
                 &&\multicolumn{3}{c|}{\texttt{semiCRF}} & \multicolumn{3}{c|}{\texttt{joint}} & \multicolumn{3}{c|}{\texttt{s2s}} & \multicolumn{3}{c|}{\texttt{PGNet}}&\multicolumn{3}{c}{\texttt{IL}}\\\midrule
    DS&Lang.  & Acc.     & ED        & F1    & Acc.     & ED        & F1       & Acc.     & ED     & F1       & Acc.     & ED       & F1       & Acc.     & ED       & F1      \\\midrule
        &EN   &\bf 52.87 &\bf  80.80 & 61.27 &    48.76 &     88.32 &\bf 64.45 &    20.34 & 232.07 &    44.35 &    44.0  &   118.53 &    59.28 &    50.99 &    88.21 &    62.47\\
    100 &GR   &    35.12 &    131.78 & 61.75 &    51.33 &    109.08 &    69.07 &    19.39 & 283.53 &    60.11 &    39.29 &   140.17 &    69.99 &\bf 51.49 &\bf 94.44 &\bf 74.29\\
        &ID   &    58.82 &     68.41 & 71.44 &    52.21 &     95.13 &    70.03 &    25.41 & 207.57 &    68.7  &    53.41 &    86.16 &    78.58 &\bf 61.14 &\bf 56.56 &\bf 79.86\\\hline
        &EN   &    56.42 &\bf  75.26 & 66.91 &    52.32 &     85.37 &    70.05 &    40.34 & 137.98 &    61.74 &    53.0  &    92.57 &    69.47 &\bf 57.26 &    79.26 &\bf 70.08\\
    200 &DE   &    36.34 &    124.36 & 65.08 &    55.27 &    102.61 &    71.22 &    41.2  & 155.71 &    72.35 &    48.82 &   109.28 &    75.39 &\bf 54.90 &\bf 88.04 &\bf 76.94\\
        &ID   &    60.96 &     62.16 & 75.51 &    57.00 &     89.34 &    72.98 &    55.58 &  97.97 &    82.62 &    67.57 &    54.97 &    85.92 &\bf 71.38 &\bf 39.89 &\bf 86.05\\\hline
        &EN   &    57.30 &     73.37 & 68.50 &    55.67 &     81.39 &    72.23 &    50.27 & 107.28 &    68.35 &    56.54 &    85.81 &    72.87 &\bf 61.08 &\bf 71.67 &\bf 73.72\\
    300 &DE   &    38.10 &    118.48 & 68.28 &    56.51 &    101.18 &    71.85 &    49.94 & 118.36 &    76.73 &    52.91 &    97.17 &    77.78 &\bf 58.82 &\bf 79.78 &\bf 79.16\\
        &ID   &    62.68 &     58.48 & 77.92 &    59.22 &     82.89 &    75.23 &    68.09 &  62.39 &    87.29 &    73.74 &    42.84 &    88.87 &\bf 76.12 &\bf 32.06 &\bf 88.92\\ \hline
        &EN   &    58.54 &     71.78 & 69.78 &    58.08 &     74.79 &    74.35 &    55.10 &  92.78 &    72.11 &    59.58 &    79.63 &    75.68 &\bf 63.14 &\bf 68.32 &\bf 76.35\\
    400 &DE   &    38.27 &    116.31 & 69.11 &    57.17 &    101.23 &    65.59 &    55.14 &  99.11 &    79.22 &    55.16 &    92.99 &    78.81 &\bf 60.37 &\bf 77.61 &\bf 80.15 \\
        &ID   &    63.27 &     56.52 & 79.39 &    63.92 &     56.12 &    79.67 &    74.18 &  48.39 &    89.73 &    75.69 &    39.29 &    89.88 &\bf 79.32 &\bf 27.31 &\bf 90.80\\\hline
        &EN   &    59.06 &     70.87 & 69.97 &    57.73 &     74.21 &    73.71 &    60.19 &  83.62 &    76.02 &    62.38 &    74.47 &    77.45 &\bf 64.88 &\bf 65.20 &\bf 77.70\\
    500 &DE   &    38.72 &    113.84 & 70.08 &    58.53 &     99.16 &    73.00 &    57.84 &  90.63 &    80.66 &    58.78 &    85.44 &    80.20 &\bf 62.37 &\bf 73.11 &\bf 80.87\\
        &ID   &    63.93 &     55.36 & 79.75 &    66.01 &     52.42 &    78.14 &    78.43 &  39.58 &    91.49 &    78.17 &    34.76 &    91.15 &\bf 81.16 &\bf 25.16 &\bf 91.80\\\hline
        &EN   &    59.79 &     69.96 & 71.05 &    59.51 &     68.27 &    73.71 &    61.72 &  77.06 &    76.73 &    63.78 &    71.02 &    78.62 &\bf 66.67 &\bf 61.46 &\bf 79.37\\
    600 &DE   &    38.76 &    113.71 & 69.94 &    59.76 &     93.21 &    74.00 &    59.46 &  87.29 &    81.03 &    59.09 &    84.32 &    80.40 &\bf 63.12 &\bf 71.11 &\bf 81.41\\
        &ID   &    63.96 &     55.27 & 79.65 &    70.56 &     50.45 &    81.92 &    80.14 &  32.93 &    92.23 &    80.43 &    30.59 &\bf 92.36 &\bf 81.62 &\bf 24.18 &    92.15\\\bottomrule
           
    \end{tabular}
    \caption{Performance of all systems for increasing training set sizes; DS=dataset size. }
    \label{tab:complete_results}
\end{table}

\begin{table*}[]
    \centering
    \setlength{\tabcolsep}{2.5pt}
    \small
    \begin{tabular}{c | c| c | c | c | c | c | c | c | c | c | c| c | c | c}
    \toprule
         &    Overseg.  & Underseg. &  Res. &  Overres. & Wrong seg.\\\midrule
English       &  0.4    &  21.3     &  12.0 &  18.8      &  71.0\\
German        &  0.0    &  21.7     &  27.9 &   9.5      &  54.9\\
Indonesian    &  11.3   &  41.5     &  11.5 &  17.0      &   63.1\\
Tepehua       &  0.4    &  50.1    &  9.2 &  31.2     &  88.8\\
Popoluca      &  13.4  &  62.4    &  3.8 &  26.7     &  91.5\\ \bottomrule

    \end{tabular}
    \caption{Error types found in the development set for the \texttt{Joint} model.}
    \label{fig:error_joint}
\end{table*}

\begin{table*}[]
    \centering
    \footnotesize
    \begin{tabular}{c|c}
                System & Link \\\hline
      \texttt{semiCRF} & \url{http://cistern.cis.lmu.de/chipmunk/} \\
      \texttt{Joint}   & \url{https://github.com/ryancotterell/treeseg} \\
      \texttt{s2s}     & \url{https://opennmt.net/} \\
      \texttt{PGNet}   & \url{https://github.com/abhishek0318/conll-sigmorphon-2018} \\
      \texttt{IL}      & \url{https://github.com/ZurichNLP/emnlp2018-imitation-learning-for-neural-morphology} \\
    \end{tabular}
    \caption{Links to all system used in this research}
    \label{tab:links}
\end{table*}

\end{document}